# Multi-Domain ABSA Conversation Dataset Generation via LLMs for Real-World Evaluation and Model Comparison

Tejul Pandit[1], Meet Raval[2], and Dhvani Upadhyay[3]

[1] Palo Alto Networks, Santa Clara, USA
[2] University of Southern California, Los Angeles, USA
[3] Dhirubhai Ambani University, Gandhinagar, India

**Abstract.** Aspect-Based Sentiment Analysis (ABSA) offers granular insights into opinions but often suffers from the scarcity of diverse, labeled datasets that reflect real-world conversational nuances. This paper presents an approach for generating synthetic ABSA data using Large Language Models (LLMs) to address this gap. We detail the generation process aimed at producing data with consistent topic and sentiment distributions across multiple domains using GPT-4o. The quality and utility of the generated data were evaluated by assessing the performance of three state-of-the-art LLMs (Gemini 1.5 Pro, Claude 3.5 Sonnet, and DeepSeek-R1) on topic and sentiment classification tasks. Our results demonstrate the effectiveness of the synthetic data, revealing distinct performance trade-offs among the models: DeepSeek-R1 showed higher precision, Gemini 1.5 Pro and Claude 3.5 Sonnet exhibited strong recall, and Gemini 1.5 Pro offered significantly faster inference. We conclude that LLM-based synthetic data generation is a viable and flexible method for creating valuable ABSA resources, facilitating research and model evaluation without reliance on limited or inaccessible real-world labeled data.

## 1 INTRODUCTION

Aspect-Based Sentiment Analysis (ABSA) has emerged as a critical task within Natural Language Processing (NLP), moving beyond coarse-grained sentiment polarity (positive/negative/neutral) to identify sentiments associated with specific topics within a text. This granular analysis provides more profound insights that are valuable for applications ranging from product review analysis and market research to social media monitoring and customer feedback understanding.

Research in ABSA has been significantly advanced by benchmark datasets, predominantly stemming from challenges like SemEval (e.g., SemEval-2014 Task 4 [1], SemEval-2015 Task 12 [2], SemEval-2016 Task 5) [3] and other efforts such as the AWARE dataset [4]. Despite their extensive use, a prominent limitation of these datasets is that the focus of these datasets is predominantly on reviews of certain items, which limits the representativeness of real-world conversations. Additionally, these datasets only capture a small subset of topics and emotions.

A significant challenge arises when applying ABSA to broader, real-world scenarios. Authentic conversations, whether from customer service logs, social media, or forums, rarely adhere to the neat structures of benchmark datasets. They often lack reliable sentiment labels, cover many topics, and seldom maintain a single tone. Since people usually convey multiple feelings about several issues in a single speech, the ABSA task is intrinsically complex but highly relevant for capturing genuine opinions. The lack of labeled data that reflects this complexity makes it challenging to create and assess reliable ABSA models that can be applied to various real-world scenarios.

This paper addresses this gap by presenting and evaluating an approach that leverages the capabilities of modern Large Language Models (LLMs) to generate diverse and domain-specific ABSA data. Based on the four main domains that we targeted, we have added our datasets [TechConV, HealthConV, FinConV, and LegalConV] to our Github repository[1]. Section 3 entails steps to produce data with regulated distributions among subjects and emotions to offer an adaptable resource for ABSA research. We evaluate the utility of this synthetic data and provide our experimental results in Section 4 by performing a comparative analysis through testing the ability of current leading LLMs (Gemini 1.5 Pro [5], Claude 3.5 Sonnet [6], DeepSeek-R1 [7]) to correctly classify the aspect-sentiment pairs within the generated text, thereby assessing both the quality of the synthetic data and the capabilities of these models on this nuanced task.

## 2 RELATED WORK

Aspect-based Sentiment Analysis (ABSA) aims to identify sentiments expressed toward specific aspects or features within a given text, providing a more granular understanding of opinions than general sentiment analysis [8, 9]. This field encompasses several subtasks, including Aspect Extraction (AE), Opinion Extraction (OE), and Aspect-Sentiment Classification (ASC) [8]. Early approaches often tackled these tasks individually, but more recent research focuses on complex end-to-end models, particularly leveraging deep learning and transformer architectures like BERT [10] and GPT [11] as discussed in 8.

The development and evaluation of ABSA models heavily rely on the availability of high-quality, diverse datasets [8]. Early datasets, such as the Citysearch New York restaurant reviews, provided initial resources for the task [12]. However, the field significantly advanced with the introduction of benchmark datasets from the SemEval competitions, including SemEval-2014 (Task 4) with restaurant and laptop reviews and SemEval-2016 (Task 5), which expanded to eight languages [9, 12]. These datasets facilitated research in Aspect Term Extraction (ATE), Aspect Term Sentiment Analysis (ATSA), Aspect Category Detection (ACD), and Aspect Category Sentiment Analysis (ACSA) [9]. Subsequent efforts led to the creation of specialized datasets such as Sentihood, which focuses on neighborhood reviews with potential multiple entities [12], and MAMS, designed to challenge sentences with multiple aspects with opposing sentiments [9, 12]. The AWARE dataset expanded the domain coverage with smartphone app reviews [12].

Despite the contributions of these datasets, several limitations have been identified [8, 9]. A significant concern is the domain skew, with a heavy concentration of research and dataset usage in the product and service review domain, particularly restaurants and electronics [8]. This lack of diversity restricts the models' capacity to be applied to other crucial fields like healthcare and education [8]. Furthermore, there is an over-reliance on a few benchmark datasets, such as those from SemEval, which are often criticized for their simplicity and lack of real-world complexity [8]. Annotation inconsistencies and the absence of standardized dataset formats also pose challenges [9].

Recognizing the predominantly English-centric nature of early ABSA datasets, the M-ABSA dataset was created to provide a large-scale parallel resource for aspect triplet extraction (TASD) across 21 languages and seven domains [13]. This effort addresses the need for multilingual resources to facilitate cross-lingual research and evaluation [13].

A critical aspect often overlooked in traditional ABSA is identifying conflict sentiment, where an element simultaneously evokes positive and negative opinions [14]. Existing ABSA research, including many approaches using Pre-trained Language Models (PLMs), often focuses on binary or ternary sentiment classification, neglecting this vital nuance [14]. The scarcity of datasets adequately annotated with conflict sentiment further exacerbates this issue [14]. To address this, some prior work has explored four-class sentiment classification, including conflict, and investigated using data augmentation techniques to overcome the data imbalance [14]. Thus, it motivates the need for methods to generate more data, particularly for underrepresented sentiment classes, to improve the robustness and accuracy of ABSA models.

---

[1] Code, data, and prompts available in Github

## 3 DATASET

This section details the process of creating a synthetically generated dataset, and evaluate its reliability by assessing various models for abstract based sentiment analysis. The objective is to enable generation-based learning and capture sentiment-oriented language patterns. We used GPT-4o [15] to construct this dataset, guaranteeing enough coverage, diversity, and scalability of different sentiment expressions. The generated dataset comprises of conversations with varying sentiments (positive/negative/neutral) across a range of topics (ref. Appendix A) that capture the nuances commonly found in natural language.

### 3.1 Dataset Generation

We use a structured, multi-stage pipeline to generate synthetic data to overcome the constraints of existing real-world datasets for ABSA task, explicitly concerning their size, diversity, domain specialization, and ability to capture complex sentiment expressions. With this method, we can produce a domain-specific, scalable, and controlled dataset. The main objective of this pipeline is to provide a vast corpus of realistic, context-rich, sentiment-aligned conversations that demonstrate a variety of sentiment dynamics and serve as a valuable foundation for learning objectives based on generational data. The process begins with selecting a domain, such as technology or healthcare, which defines the thematic boundaries and relevant vocabulary. Within each domain, we prompt GPT-4o to generate 10 distinct types of conversations that reflect common interaction patterns, detailed in Section 3.1.1. Subsequently, we use GPT-4o to obtain 20 domain-relevant topics across these conversation types to guide the focus areas for the generated dialogue. To ensure comprehensive sentiment coverage, each topic is paired with a sentiment label (positive, negative, or neutral) using a sophisticated selection mechanism described in Section 3.1.2. Finally, using a composite prompt that incorporates the selected domain, a randomly chosen conversation type, and a topic-sentiment pair, we generate realistic, multi-turn conversations that convey the intended sentiment in natural language. Section 3.1.3 covers the details regarding the prompt for conversation generation. The resulting dataset is diverse and sentiment-controlled making it especially suited for building and evaluating robust models for ABSA.

#### 3.1.1 Conversation Scenario Creation:
We start by narrowing our research of data generation to four sets of domains entailing Technology, Healthcare, Finance, and Legal. The next crucial stage in our data-generation pipeline was to discover and choose discussion scenarios within each of the four domains that had been selected. Therefore, identifying suitable discussion kinds was crucial to our objective of producing high-quality, synthetic, multi-turn dialogue data.

To address this, we leveraged GPT-4o to explore the space of realistic conversation settings. For each domain, we designed prompts asking the LLM to enumerate ten distinct conversation scenarios that typically involve multiple participants to mimic a multi-turn conversation. These prompts were thoughtfully designed to enable the model to have creative latitude in its outputs while eliciting scenarios based on routine operational or professional operations. We ensured the resulting recommendations were varied, organic, and contextually based by letting the LLM suggest conversation scenarios naturally rather than enforcing rigid structural limits.

The LLM-generated conversation captured formal and informal interaction patterns and mirrored real-world communicative contexts such as planning meetings, evaluations, negotiations, consultations, and cooperative talks. Creating a strong contextual framework for dialogue production that would facilitate realistic role-based communication, genuine interaction dynamics, and cogent discourse flow was our goal in this phase.

This process yielded 40 distinct conversation scenarios (10 per domain), forming the backbone of our dataset's structural diversity. These types serve as high-level templates, guiding the subsequent topic selection and dialogue synthesis stages.

**3.1.2 Topic Selection:** Having created the conversation scenarios, the next step would be to identify the thematic content of each dialogue. For every type of conversation, we again utilized the GPT-4o to generate a list of 10 topics that would commonly arise in that particular communicative context. This led to a total of 100 topics (10 topics $\times$ 10 conversation scenarios) for each domain, spanning a wide thematic range.

However, we found that the topics generated for a conversation scenario had considerable similarity with the topics of the other conversation scenarios. This led us to reduce the total topics to 20 for each domain. This reduction was carried out by again leveraging GPT-4o and prompting it to consider all the 100 topics and return a new list of 20 topics which highly representative of their respective domains, but are also sufficiently generic to support diverse dialogue structures. We repeated the same approach for all the domains included in our research.

**3.1.3 Conversation Generation:** To replicate a real-world communication, each synthetic conversation generated through our approach was created such that there are multiple topics that are being discussed in the conversation. For each of the topics being discussed, a distinct sentiment is assigned with the restriction that all the topics cannot have the same sentiment. For the purpose of our research, each conversation sample was designed to be generated using a random conversation type for the domain and either two or three different topics.

We developed a topic selection algorithm to ensure a near-symmetric distribution of topics and sentiments. The pseudo-code for the implemented algorithm is presented in Algorithm 1. This prompting strategy enabled precise control over the distribution of topics within each conversation. For a total number of topics n, iterated m times, the time complexity of Algorithm 1 is $O(mn)$.

In total, 12,000 synthetic conversations were generated—3,000 for each domain of interest. These conversations were subsequently filtered to eliminate highly similar samples, thus improving the variability of the dataset while maintaining a manageable data volume.

**Algorithm 1** Conversation Generation Algorithm
---
1: **Input:** topics = list of n topics
2: **Input:** m = number of outputs
3: **Initialize:**
4:     topic_frequency ← $\frac{m}{n}$
5:     assigned_outputs ← empty list
6:     topic_usage ← map each topic → 0
7:     sentiment_proportions ← proportions for positive, negative and neutral sentiments
8:     conversations ← empty list
9: **for** each topic in topics **do**
10:     **for** $i = 1$ to topic_frequency **do**
11:         **Step 1:** Select the number of topics to use (2 or 3)
12:         topics_to_use_count ← randomly select 2 or 3
13:         **Step 2:** Initialize the selected topics
14:         selected_topics ← [topic]
15:         **Step 3:** Get additional topics to complete the set
16:         additional_topics ← getAdditionalTopics(topics_to_use_count - 1, topic, topics)
17:         **Step 4:** Assign sentiments to the selected topics based on sentiment proportions
18:         sentiments ← getRandomSentiments(topics_to_use_count)
19:         **Step 5:** If all sentiments are the same, restart this iteration
20:         **if** all sentiments are the same **then**
21:             redo this iteration
22:         **end if**
23:         **Step 6:** Select a random conversation scenario
24:         conversation_scenario ← getRandomScenario()
25:         **Step 7:** Generate prompt based on selected topics and sentiments
26:         prompt ← generatePrompt(selected_topics, sentiments, conversation_scenario)
27:         **Step 8:** Add generated prompt to conversations list
28:         Add prompt to conversations
29:     **end for**
30: **end for**
31: **Return:** conversations

## 3.2 Data Filtering and Validation

We implemented a multi-step data filtering and validation pipeline grounded in semantic similarity and filename-encoded sentiment annotations to construct a high-quality dataset for aspect-based sentiment analysis in each domain. An organized directory containing .txt files containing aspect indices and the sentiment labels that corresponded to them was used to extract raw conversation data. These indices were mapped to a predefined list of twenty topics, creating structured JSON-like annotations for each conversation.

We incorporated an LLM as a judge to assess the semantic integrity of the generated data. Specifically, we used Gemini 1.5 Pro to verify whether each conversation accurately reflected the assigned topics and their corresponding sentiment labels. Only those samples that the model validated as consistent and faithful to the prompt structure were retained for further processing. Using an LLM as a judge model also helps to mimic the two-annotator approach as discussed in 9, which is the most frequently used annotation procedure for ABSA dataset generation.

For semantic validation, we used Sentence-BERT (SBERT) [16] to generate dense vector embeddings for each conversation. We calculated cosine similarity scores between each conversation and every other entry in the dataset rather than using a strict cutoff level. Each conversation was then connected to the collection of all other conversations with a similarity score $\geqslant 0.8$ in our created similarity map. This mapping represented each conversation's contextual closeness within the dataset.

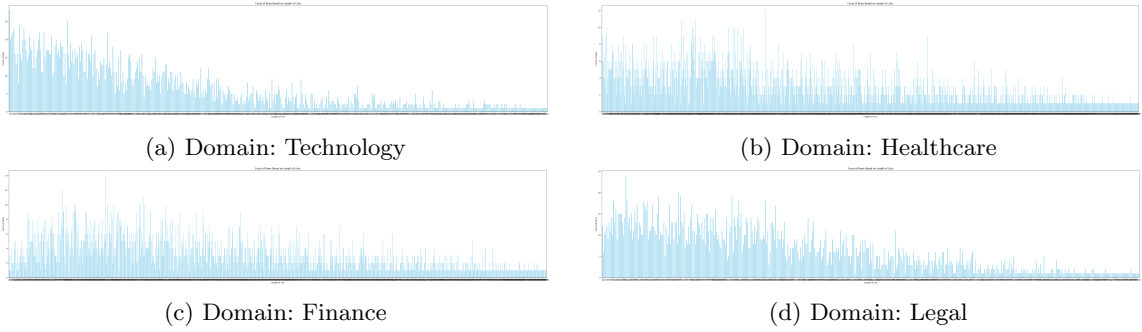

Fig. 1: Conversation Count by List Length

To prioritize data requiring human inspection and cleanup, we sorted the dataset in descending order based on the length of each similarity list (i.e., number of highly similar neighbors). Conversations with long similarity lists typically exhibit thematic repetition or generic phrasing, making them less informative. By analyzing the distribution of similarity list lengths across all conversations, we identified the top 33% of conversations with the highest number of neighbors as candidates for manual review and pruning. This approach allowed us to systematically surface potentially redundant or noisy entries while preserving diversity in sentiment expression across aspect-specific dimensions. Fig. 1 captures the distribution of the count of conversations based on the length of other matching conversations for each domain.

This process yielded a dataset in which each remaining conversation is aspect-annotated and semantically distinct, making it suitable for downstream tasks requiring contrastive sentiment learning in multi-aspect contexts.

### 3.3 Dataset Statistics

The final dataset consists of artificial dialogues produced in several domains, each intended to replicate authentic sentiment dynamics and interaction patterns. To guarantee coverage and diversity, the dataset contains a balanced distribution of discussion topics and kinds within each area, including neutral, negative, and positive attitudes. Table 1 summarizes key statistics for each domain, including the total number of conversations, average word count per conversation, and the distribution of sentiment classes.

Table 1: Dataset statistics post filtering across different domains.

| Domain | Total Conversations | Avg. Words | Sentiment Distribution (Pos/Neg/Neu) |
|---|---|---|---|
| Technology | 2,036 | 176 | 33%/32%/35% |
| Healthcare | 2,082 | 157 | 33%/34%/34% |
| Finance | 2,011 | 169 | 33%/32%/34% |
| Legal | 2,166 | 152 | 34%/33%/32% |

To ensure structural consistency, we examined the frequency distribution of sentiment-class and topic occurrences in each of the four domains in addition to raw statistics. Based on the fundamental design explained in Section 3.1.3, topic and sentiment representations are relatively uniform and well-balanced. Appendix A provides a visual representation indicating the similar distribution across all topics in each of the domains used for our paper. This uniformity was maintained despite the filtering process discussed in Section 3.2. This consistency for training and assessing reliable ABSA models across domains demonstrates that the data creation pipeline effectively maintained coverage across several aspect categories while preserving parity in sentiment expression.

## 4 RESULTS & OBSERVATIONS

This section presents the evaluation results of the synthetic data generated. We assessed the quality of the generated data by evaluating the performance of three prominent LLMs – Gemini 1.5 Pro, Claude 3.5 Sonnet, and DeepSeek-R1 – on topic and sentiment classification tasks.

### 4.1 Metric Calculation

We employed a suite of standard classification metrics to comprehensively evaluate the models' performance on the synthetic data for both topic (aspect) and sentiment classification. Let $C$ be the set of all classes and $c$ represents each individual class.

#### 4.1.1 Aggregated Metrics

– **Subset Accuracy:** Measures the proportion of samples for which the predicted set of labels ($\hat{y}^{(i)}$) matches the true set ($y^{(i)}$).

$$\text{Subset Accuracy} = \frac{1}{N} \sum_{i=1}^{N} \left[ \hat{y}^{(i)} = y^{(i)} \right]$$

– **Micro-Averaged Metrics:** Calculates metrics globally by summing the individual true positives (TP), false positives (FP), and false negatives (FN) across all classes before applying the standard formulas. It weights each instance equally.
– **Macro-Averaged Metrics:** Calculates each class's metric (e.g., Precision) independently and then takes the unweighted arithmetic mean of these scores. It treats all classes equally.
– **Weighted-Averaged Metrics:** Calculates the metric for each class independently and then takes the weighted average, where each class's score is weighted by its support (the number of true instances for that class, $\text{Support}_c = \text{TP}_c + \text{FN}_c$). This accounts for class imbalance. Let $N$ be the total number of cases ($N = \sum \text{Support}_c$).

**4.1.2 Performance Results** The performance of each LLM on the generated synthetic data across the four domains is summarized in Tables 2, 3, 4, and 5. Each table presents the domain-based metric results as discussed in Section 4.1.1. Precision, Recall, and F-1 scores are calculated on average. Additionally, the bold numbers in the table highlight the highest value obtained for the specific metric. Detailed analysis and observations on the results obtained are presented in Section 4.1.3

Table 2: Model Comparison on Domain: Technology

| Metric | Gemini | Claude | Deepseek |
|--------|--------|--------|----------|

|                | Prec | Rec  | F1   | Prec | Rec  | F1   | Prec | Rec  | F1   |
|----------------|------|------|------|------|------|------|------|------|------|
| **Micro Avg**    | 0.71 | **0.98** | 0.82 | 0.71 | **0.98** | 0.82 | **0.80** | 0.96 | **0.87** |
| **Macro Avg**    | 0.79 | **0.98** | 0.86 | 0.76 | **0.98** | 0.84 | **0.82** | 0.96 | **0.87** |
| **Weighted Avg** | 0.79 | **0.98** | 0.86 | 0.77 | **0.98** | 0.85 | **0.83** | 0.96 | **0.88** |
| **Samples Avg**  | 0.74 | 0.98 | 0.83 | 0.74 | **0.99** | 0.83 | **0.84** | 0.96 | **0.88** |

Table 3: Model Comparison on Domain: Healthcare

| Metric | Gemini | | | Claude | | | Deepseek | | |
|--------|--------|--|--|--------|--|--|----------|--|--|
|        | Prec | Rec | F1 | Prec | Rec | F1 | Prec | Rec | F1 |
| **Micro Avg**    | 0.62 | **0.98** | **0.76** | 0.63 | 0.86 | 0.73 | **0.70** | 0.84 | **0.76** |
| **Macro Avg**    | 0.67 | **0.98** | **0.78** | 0.66 | 0.86 | 0.75 | **0.72** | 0.83 | 0.77 |
| **Weighted Avg** | 0.68 | **0.98** | **0.79** | 0.67 | 0.86 | 0.75 | **0.75** | 0.84 | **0.79** |
| **Samples Avg**  | 0.67 | **0.98** | 0.78 | 0.67 | 0.86 | 0.75 | **0.75** | 0.84 | **0.79** |

Table 4: Model Comparison on Domain: Finance

| Metric | Gemini | | | Claude | | | Deepseek | | |
|--------|--------|--|--|--------|--|--|----------|--|--|
|        | Prec | Rec | F1 | Prec | Rec | F1 | Prec | Rec | F1 |
| **Micro Avg**    | 0.65 | 0.99 | 0.79 | 0.65 | **1.00** | 0.78 | **0.77** | 0.98 | **0.86** |
| **Macro Avg**    | 0.75 | 0.99 | 0.82 | 0.73 | **1.00** | 0.82 | **0.81** | 0.98 | **0.87** |
| **Weighted Avg** | 0.79 | 0.99 | 0.85 | 0.77 | **1.00** | 0.84 | **0.84** | 0.98 | **0.89** |
| **Samples Avg**  | 0.70 | 0.99 | 0.80 | 0.70 | **1.00** | 0.80 | **0.81** | 0.98 | **0.88** |

Table 5: Model Comparison on Domain: Legal

| Metric | Gemini | | | Claude | | | Deepseek | | |
|--------|--------|--|--|--------|--|--|----------|--|--|
|        | Prec | Rec | F1 | Prec | Rec | F1 | Prec | Rec | F1 |
| **Micro Avg**    | 0.72 | 0.96 | 0.82 | 0.72 | **0.98** | 0.83 | **0.76** | 0.93 | **0.84** |
| **Macro Avg**    | 0.76 | 0.96 | 0.84 | 0.78 | **0.98** | **0.86** | **0.82** | 0.93 | **0.86** |
| **Weighted Avg** | 0.77 | 0.96 | 0.85 | 0.79 | **0.98** | **0.86** | **0.82** | 0.93 | **0.86** |
| **Samples Avg**  | 0.77 | 0.96 | 0.84 | 0.76 | **0.98** | 0.84 | **0.80** | 0.93 | **0.85** |

We similarly assessed the sentiment classification as represented in Figure 3. These charts visually illustrate an overall similarity in sentiment performance, with the exception noted in the Healthcare domain. The difference in performance for a specific domain could be attributed to the disparity in training of each of these LLMs or certain anomaly introduced during label generation.

**4.1.3 Analysis & Interpretation** The evaluation results reveal several key observations regarding the utility of the generated synthetic data and the comparative performance of the LLMs.

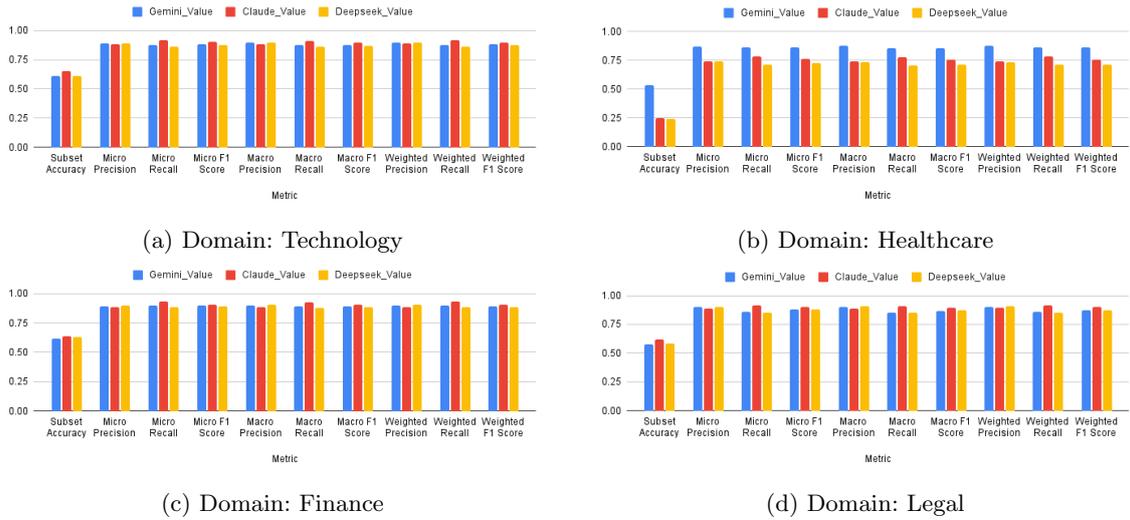

(a) Domain: Technology

(b) Domain: Healthcare

(c) Domain: Finance

(d) Domain: Legal

Fig. 2: Metrics for Sentiment Classification Across LLMs

Across all four domains, DeepSeek-R1 consistently demonstrated superior performance in Precision and, consequently, F1-score for topic classification tasks (refer to Tables 2 - 5). While DeepSeek-R1 led in precision, Gemini 1.5 Pro and Claude 3.5 Sonnet exhibited competitive Recall scores, often surpassing DeepSeek-R1. High recall implies that these models were better at identifying the most relevant topic instances in the synthetic data, even if they made more incorrect predictions (lower precision).

However, a clear trade-off between classification performance and inference speed emerges. Gemini 1.5 Pro offered significantly faster response generation (∼2s) compared to Claude 3.5 Sonnet (∼7s) and DeepSeek-R1 (∼15s). Observing precision v/s latency highlights a critical consideration for practical ABSA deployments: applications requiring real-time analysis might favor Gemini despite slightly lower accuracy. In contrast, applications prioritizing maximum accuracy might opt for DeepSeek-R1, accepting the higher latency.

The generally strong performance across all three state-of-the-art models suggests that the synthetic data generation process successfully created relevant and challenging examples for the ABSA task. The differing strengths of the models (precision vs. recall) indicate that the synthetic data likely contains a mix of clear-cut and more subtle or ambiguous examples, reflecting complexities often found in real-world data. The ability of models like Gemini and Claude to achieve high recall suggests good coverage of aspect/sentiment expressions within the synthetic dataset.

## 5   CONCLUSION

This study demonstrated the viability of using our approach to synthetically generate data for evaluating state-of-the-art LLMs on complex ABSA tasks across multiple domains. Our findings indicate that the generated dataset contains subtle variations and complexity that allow us to meaningfully compare how well different LLMs perform, highlighting their individual strengths and limitations.

Synthetically generating data for a specific task offers significant practical benefits. We successfully ensured a consistent distribution of data across the targeted topics and sentiments, providing a balanced dataset for evaluation. Furthermore, the prompt-based approach also facilitates expansion of the dataset; for instance, this opens a channel to experiment with additional sentiments and evaluate the performance. Researchers may easily modify prompts to generate data for other individuals or moods, circumventing any potential challenges of acquiring labeled data from the real world.

In conclusion, our synthetic data generation approach presents a powerful and flexible tool for advancing ABSA research. It enables robust model evaluation and comparison while offering scalability for future investigations into diverse domains and aspect/sentiment categories. The observed performance variations also underscore the need for careful model selection based on the intended application's specific accuracy, recall, and latency requirements.

## A  Topic Distribution

Bar chart representation of all topics across the four domains post data cleaning and filtering process.

(a) Domain: Technology

(b) Domain: Healthcare

(c) Domain: Finance

(d) Domain: Legal

Fig. 3: Topic Distribution Charts Across Domains